# FACE RECOGNITION USING PCA INTEGRATED WITH DELAUNAY TRIANGULATION

Kavan Adeshara and Vinayak Elangovan

Division of Science and Engineering, Penn State Abington, PA, USA
`kxa5317@psu.edu`
Division of Science and Engineering, Penn State Abington, PA, USA
`vue9@psu.edu`

## *ABSTRACT*

*Face Recognition is most used for biometric user authentication that identifies a user based on his or her facial features. The system is in high demand, as it is used by many businesses and employed in many devices such as smartphones and surveillance cameras. However, one frequent problem that is still observed in this user-verification method is its accuracy rate. Numerous approaches and algorithms have been experimented to improve the stated flaw of the system. This research develops one such algorithm that utilizes a combination of two different approaches. Using the concepts from Linear Algebra and computational geometry, the research examines the integration of Principal Component Analysis with Delaunay Triangulation; the method triangulates a set of face landmark points and obtains eigenfaces of the provided images. It compares the algorithm with traditional PCA and discusses the inclusion of different face landmark points to deliver an effective recognition rate.*

## 1. INTRODUCTION

### 1.1. Face Recognition and Principal Component Analysis

Face recognition has always been a topic of keen interest for computer vision researchers. It is extensively studied to advance the system's efficiency and minimize errors. Notably, PCA is one of the prominent approaches applied by the researchers to develop user-authentication system; it allows a user to train programs in distinguishing individual faces by providing a set of training and testing data. Using mathematical functions such as Singular Value Decomposition (SVD), the approach obtains eigenvalues and eigenvectors to project an image onto the eigenspace. The projected images, called Eigenfaces, helps to determine the Euclidean distances between the testing and training images, recognizing the training image with the least distance.

According to studies [1] and [2], employing PCA generates fewer errors when comprehensive training data is provided, such as variations in face illumination, expressions, and angles. However, when insufficient data are trained on, this approach fails to retain an accurate recognition rate. In most applications, all imagery data are set to be in a grayscale format with a preferred dimension. Colored images and images with varying dimensions can often be tedious to process leading to increase in computation time.

### 1.2. Alternate Variants of Principal Component Analysis

To mitigate the limitation of traditional PCA based face recognition, several alternative versions of PCA have been proposed in the past decade. For instance, Article [3] proposes a modular PCA algorithm that divides an image into smaller sub-images which are least affected by variable changes such as a change in pose and illumination. The algorithm implements PCA on these unaltered sub-images. Using the Yale Face, ORL, and UMIST datasets [4], the study concludes an improved efficacy of their proposed algorithm, unaffected by face pose, lighting, and face tilt.

Article [5] examines the implementation of a composite Kernel PCA algorithm which successfully deals with a large training database by outlining the data into high dimension vector space with non-linear transformation. Using the ORL and FERET [4] face database, the algorithm outputs better processing and recognition rates of above 90% for a large sample size but falls short on effective time consumption.

### 1.3. Other Notable Techniques

Besides PCA, different unique techniques have been executed for the development of a robust face recognition system. In [6], the implementation of Delaunay Triangulation for face verification is discussed. The approach sets the image in a triangular plane where the triangulation would distribute the entirety of the face into normalized triangles. A squared difference between the training and testing image's triangular area is calculated to determine the closest recognized image in the database. The article experiments the technique with 15 frontal, 256 x 256 images, observing an improved recognition rate with illumination and pose change.

Article [7] proposes a novel technique that uses a Radial basis function neural network to acquire the centers of hidden neuron layers for face recognition. From the experimental results using Yale Face, ORL, AR, and LF dataset [4], it shows a clear advantage of combining the firefly algorithm with the neural network that significantly optimizes the feature selection and speeds up the convergence rate. Article [8] proposes an algorithm that utilizes a combination of three different algorithms: Wavelet Transformation, Local Linear Embedding, and Support Vector Machines. The proposed algorithm breaks the face image into four components using wavelet transformation, then uses local linear embedding to analyze the key features of the four components, after which a weighted fusion is determined to perform face recognition. Lastly, SVM will be utilized to train the eigenvectors of the face data and the face classifier class. The algorithm is tested on three different image dataset and the algorithm yields an improved accuracy rate among all the mentioned datasets. However, one key area of improvement needed is to determine a reasonable weight among different face image weights. In the article [9], Gabor filter is used to obtain Gabor amplitude of face images after which Uniform Local Binary Histogram is obtained. Then, a dictionary is implemented using Fisherion criterion and an image is classified using spare representation coding. Using the Yale B and AR image datasets, the amalgamtions of all these algorithms yielded an improved recognition rate in variable image environments such as change in lighting, illumination etc.

### 1.4. PCA integrated with different techniques

Some studies combine PCA with different novel techniques to achieve a better computational and accuracy rate. Namely, researchers at [10] and [11] integrates PCA with Delaunay Triangulation, Fisherface algorithm, and Convolutional Neural Network respectively to provide increased accuracy with minimal time needed for computation.

In [10], a face model is derived by implementing a synthesis of PCA, Delaunay Triangulation, and Fisher face algorithm. It procures the shape and the texture of the face by acquiring key information from both the triangulation of face landmarks and the Fisher faces using PCA. It applies a collection of 22 points for face landmarks. Colored images from the AR database are used for testing the proposed algorithm. It yields an average of 95% and above when a large sample is trained for biometric verification.

In [11], a combination of Color 2D-PCA with Convolutional Neural Network (CNN) is developed which integrates feature extraction of CNN and 2D-PCA into one decision-level fusion. The algorithm uses the CNN model for depth features, and Color 2D-PCA for detailed image color and spatial information. The experiments, conducted using LFW and FRGC database, show a reduced training time and increased accuracy of above 90% for various image noise levels.

## 1.5. Integration of PCA with Delaunay Triangulation

This paper investigates the combination of PCA with Delaunay Triangulation to accomplish an enhancement in the face recognition system. The amalgamation of the two approaches demonstrates a slight improvement in the recognition rate when a certain number of landmarks are employed.

The program acquires a set of training images and converts the given images into a triangular mesh of key face landmark points while also projecting all the images onto the eigenspace. The information is stored in the database by conjoining both the Euclidean distances and the average relative area of the following images, weighing both the values equally. The research draws a comparison between the traditional PCA and three different variations of landmarks in the investigated approach to see which approach and variation produces an improved recognition rate.

The following sections are organized as follows: Delaunay triangulation implementation, Principal Component Analysis, integration of PCA with Delaunay triangulation, experiment and results, and conclusion.

## 2. DELAUNAY TRIANGULATION IMPLEMENTATION

### 2.1. Introduction and Application

In computational geometry, Delaunay Triangulation refers to the triangulation of a set of distinct points such that no point in the set is inside the circumcircle of any triangle. It maximizes the minimum of angles of the triangles. The triangulation does not occur when the given points are in the same line. The condition that should be met when performing the triangulation, often referred to as the 'Delaunay condition', is that the circumcircle of any triangle should have empty interiors.

Figure 1 describes the steps for conducting Delaunay triangulation where a set of point is first plotted. A triangle is drawn using any of the three points, and a circumcircle of the triangle is drawn to check the Delaunay condition. After the condition is satisfied, a triangle is drawn including the leftover point. Once Delaunay condition is satisfied for the second triangle, this mathematical function provides a final triangulation of four points.

In computer vision, Delaunay Triangulation generates a triangular mesh of key face landmark points such as the eyes, nose, mouth, and the shape of the face. Its application underlies within face swapping, face effects seen in social media app filters, and emotion detection [12] based on the triangulation of different expressions. For face recognition, the approach can provide key information about the change in average relative area of the triangulation when different expressions are posed.

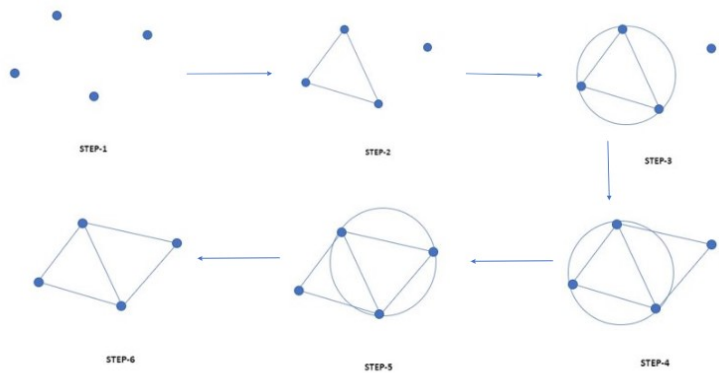

Figure 1: Graphical Representation of the concept

## 2.2. Developed Step-by-Step Implementation of Delaunay Triangulation

- ➢ Establish a training dataset for Delaunay.
- ➢ Convert all the images in the training dataset into vector to get pixel value information.
- ➢ Divide all the image vectors by 255 to simplify the mathematical calculations.
- ➢ Find face landmarks from the vectors. There are various techniques to find face landmark points. Some techniques [13] use neural network that train a classifier by feeding numerous images with manual hand drawn face landmarks on numerous faces whereas some techniques find face landmarks by finding a face that minimizes the deviation with its mean shape [14].

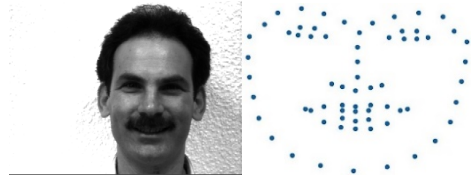

Figure 2: Sample image illustrated with 68 face landmarks [15]

- ➢ Perform Delaunay Triangulation on the face landmark points. A plethora of different algorithms exists for the triangulation. In this case, the Sweep hull algorithm is used where a sweep-hull is created in an ordered manner by looping over a set of two-dimensional points, connecting the triangles in the visible area of the convex hull.

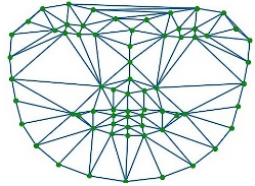

Figure 3: Delaunay Triangulation of Figure 2

- ➢ Obtain the set of vertices of all the triangles in the triangulation. This can be done by obtaining the simplices of the triangulation.
- ➢ Find edge lengths of all the triangles. The equation used for finding an edge length is as follows:

$$L = \sqrt{(x_1 - x_2)^2 + (y_1 - y_2)^2}$$

*where L is the edge length*

- ➢ Obtain areas of all the triangles from the previously calculated edge lengths.

$$S = \frac{L_1 + L_2 + L_3}{2}$$

$$A = \sqrt{S(S - L_1)(S - L_2)(S - L_3)}$$

*where S is the semi perimeter, and A is the area*

- ➢ Calculate the relative area of each triangular area.

$$RA = \frac{A_i}{A_{max}}$$

*where RA is the relative area, $A_i$ is each triangular area, and $A_{max}$ is the largest area in the triangulation*

- ➢ Obtain the average relative area of all the relative areas

$$RA_{avg} = \frac{RA_1 + RA_2 + RA_3 + \cdots + RA_e}{N}$$

*where N is the total number of average relative areas in the triangulation*

➢ Perform Delaunay Triangulation on a test image and acquire its average relative area. Save the average relative area of all the image in the database for error estimation and face comparison.

## 3. PRINCIPAL COMPONENT ANALYSIS

### 3.1. Introduction and Application

Principal Component Analysis (PCA) is a dimensionality reduction method that reduces the dimensionality of a large dataset by transforming the given dataset into a smaller one that contains enough information to represent the large dataset. It is used in many fields such as medical and technological field to analyze a large chunk of data by extracting only the vital information from the data and discounting the surplus data.

In face recognition, the method is used to reduce the dimensions of the pixel value of the image in a manner that, after the reduction, the compressed image should have ample information to represent the majority of the unique features of the image.

### 3.2. Step by Step Implementation of PCA

The following section elaborates the steps followed in developing PCA.

➢ Establish a training dataset to perform Principal Component Analysis.

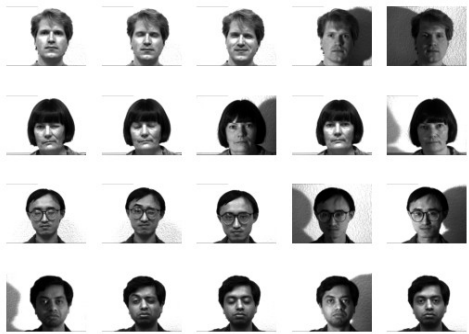

Figure 4: Images taken from Yale Face Dataset

➢ Convert all the images into vectors.
➢ Find the mean of all the image vectors.

$M = \frac{\sum_{i=1}^{n} x_i}{n}$ where M is the mean image, x is the individual image, and n is the total image

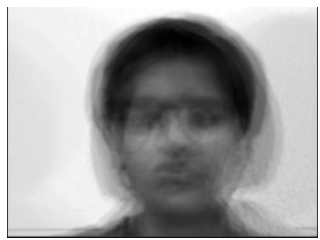

Figure 5: Mean face of all images from Figure 4

- Subtract the mean image vector from each individual image vector in the dataset.

  $I = M - x_i$

  *where I is the subtracted image vector*

- Perform Singular Value Decomposition to calculate the eigen vectors and eigen values.

  $SVD = U \sum V^T$

  *where U and $V^T$ are orthogonal matrix of eigenvector, and $\sum$ is the diagonal matrix of eigenvalues*

- For k best eigen values, reconstruct the image vectors and project them onto the eigen space. In this research, k is set to 25 for the projection of image vectors onto the eigen space.

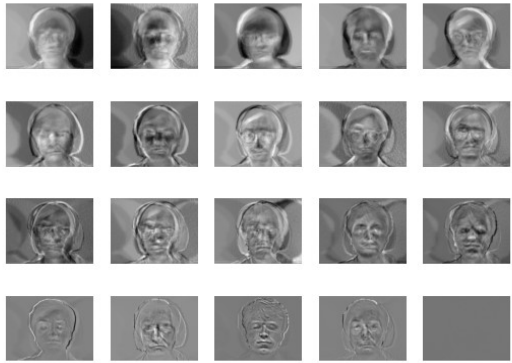

Figure 6: Eigenfaces of all images from Figure 4

- Store the information of the eigenfaces in the training database.
- Establish a testing dataset.
- Perform the same steps by converting all the testing images into image vector.
- Subtract the testing image vectors from the training mean image vector.
- Calculate the eigen values and eigenvectors of the subtracted test image vectors.

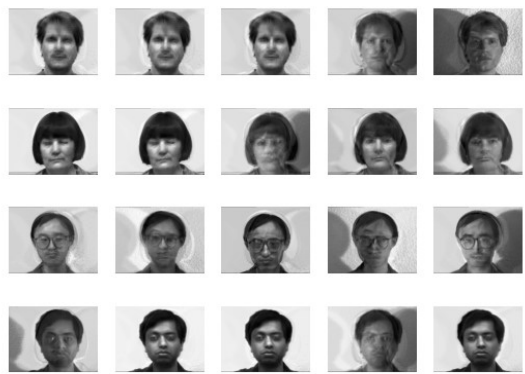

Figure 7: Reconstructed images for k = 9

- Project the images onto the eigenspace to find the closest image associated with the testing image

- Calculate the Euclidean distances of the testing image in the eigenspace with all training images and recognize the image with the least distance

## 4. INTEGRATION OF PCA WITH DELAUNAY TRIANGULATION

The following section discusses the step by step integration of PCA with Delaunay Triangulation.

- Perform Delaunay Triangulation on a set of test images and acquire their average relative areas.
- Calculate the difference between test and training images' average relative area.

$$D = \sqrt{(Tt_{avg} - Tn_{avg})^2}$$

where D is the positive difference, and $Tt_{avg}$ and $Tn_{avg}$ ar

e the average relative areas for test and train image

- Add the Euclidian Distance from PCA and difference from Delaunay Triangulation to acquire the image with least result value.

$$RV = ED + \frac{D}{0.001}$$

where RV is the resultant value, ED is the Euclidian distance and D is the difference from the triangulation

## 5. EXPERIMENTS AND RESULTS

Three experiments were conducted for testing the traditional PCA with DT (Delaunay Triangulation) integrated PCA. For each experiment, different amount of training and testing images are used. Furthermore, different number of landmark points such as 68, 79 and 194 landmarks as shown in Figure 8, are also used for the triangulation to further test the efficiency of each landmark combination to determine which landmarks work best with the discussed integration.

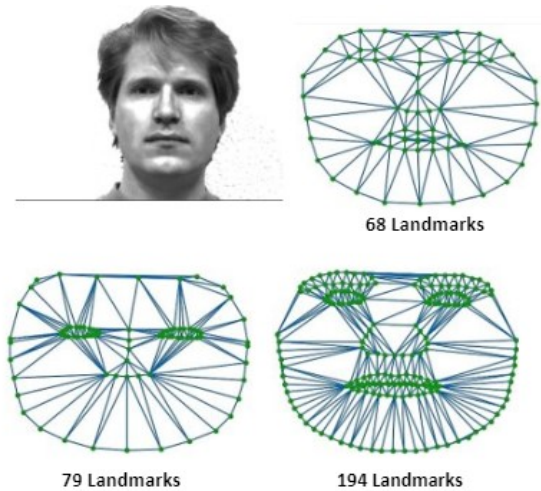

Figure 8: Triangulation of different landmarks

### 5.1. Images used for the experiment

For all experiments, 135 frontal, 320 x 243 images from the Yale Face dataset are used. Each image is in grayscale format. The dataset contains 9 total variations per individual and the variations are as follows – sleepy, happy, sad, wink, glasses, surprised, left-light, right-light, and normal. The different conditions in the dataset help assess the accuracy of the two approaches being analyzed.

## 5.2. Variation in Landmarks

The paper tests three different variation in landmarks: 68, 79, and 194 landmarks [Figure 8]. The 68 landmark contains the basic shape of the eyes, nose, and the mouth. The 79 landmark contains only eyes, nose and overall shape of the face. This landmark point help in differentiating individuals if they wear a mask, but is less helpful when comparing faces without mask because of discounting width of the mouth and the location of eyebrows. The 194 landmarks contain detailed landmark points of the eyes, nose, eyebrows, and lips. This can help differentiate other individuals from the depth in eyebrows, distance of the eyes from the nose and such.

## 5.3. Experiment 1

In experiment 1, 105 images are trained. 7 variant images of each 15 individuals were considered. The remaining images are used for testing. From the results, the recognition rates are improved for DT integrated PCA using 68, 79, and 194 face landmarks.

## 5.4. Experiment 2

In experiment 2, 75 images are trained on, from which 5 images are variants. The rest of the images are utilized as test images. The recognition rate for DT-PCA 68-L, DT-PCA 194-L is higher comparatively higher than traditional PCA. DT-PCA 194-L, especially, yields a significant improvement out of all the three different variation of landmarks in the integration.

## 5.5. Experiment 3

In experiment 3, 45 images are trained on, from which 3 image are variants. 90 images are tested. It can be observed that the accuracy of DT-PCA 79-L is less than the traditional PCA. This is because the landmark does not account for expression change and individuals that closely resemble each other in terms of geometrical configuration. This shows that the recognition rate of this landmark combination tends to decrease as less training dataset is provided and therefore, the combination is inefficient for use in the current integration.

Table 1: Percent accuracy for traditional PCA and variant of PCA with DT

|  | Traditional PCA | PCA with Delaunay Triangulation | | |
| --- | --- | --- | --- | --- |
|  |  | 68-L | 79-L | 194-L |
| **Train – 105** **Test – 30** | 86.7 % | 93.3 % | 90.0 % | 95.6 % |
| **Train – 75** **Test – 60** | 85.0 % | 88.3 % | 86.7 % | 91.6 % |
| **Train – 45** **Test – 90** | 82.2 % | 87.8 % | 81.1 % | 90.0 % |

On the other hand, DT-PCA 194-L has retained a better percent accuracy compared to the traditional PCA and all the combination of other landmarks. Thus, from the above experiments, the DT integrated PCA efficiently works with 194 face landmark points as it shows considerable improvement in the accuracy rate compared to the traditional PCA.

## 6. CONCLUSIONS

Principal Component Analysis can help build a robust face recognition system if it is modified or used with another approach. The algorithm has the level of versatility that enables it to be used with different techniques. This research shows one such example of the combination as it discusses the slight improvement in recognition rates when a DT integrated PCA is utilized. Moreover, the results can be improved further if appropriate face landmarks are used in the integration as seen in the results above. It can be used in an attendance marking system which could clock in and clock out employees or mark a student present or absent for a class. For future work, the face landmark detection could be fine-tuned to detect the very intricate edges of the face to give a detailed statistical view of the face and to acquire a more precise triangulated data.

However, there is a trade-off in using face recognition of any approach because factors like image clarity and certain face angle are always going to remain an issue. The system cannot efficiently recognize if there is an overhead or side view of the face and there is always a factor of a person undergoing various sort of surgery that could completely change the look of the face., causing the system to not authenticate the face.

## ACKNOWLEDGMENTS

This work was funded by the Multi Campus Research Experience for Undergraduates (MCREU) program, Penn State University.

**Authors**

Kavan Adeshara is a current sophomore at Penn State University, majoring in Computer Science. His research interests include artificial intelligence, computer vision and software development. He has participated in multiple projects including data analysis of global temperature, face recognition, college major recommendation applications etc.

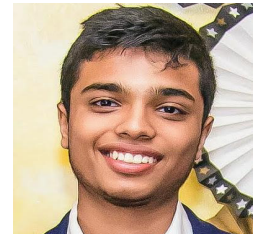

Dr. Vinayak Elangovan is an Assistant Professor of Computer Science at Penn State Abington.  His research interest includes computer vision, machine vision, multi-sensor data fusion and activity sequence analysis with keen interest in software applications development and database management. He has worked on number of funded projects related to Department of Defense and Department of Homeland Security applications.

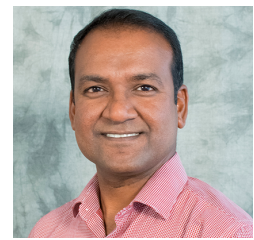